\documentclass{article}

\usepackage[preprint]{neurips_2026}
\usepackage{xcolor}
\usepackage{booktabs}
\usepackage{xcolor}
\usepackage{array}

\newcommand{\drop}[1]{\hspace{2pt}\textcolor{red!80!black}{\footnotesize($\downarrow$#1\%)}}

\usepackage[utf8]{inputenc} 
\usepackage[T1]{fontenc}    
\usepackage{hyperref}       
\usepackage{url}            
\usepackage{booktabs}       
\usepackage{amsfonts}       
\usepackage{nicefrac}       
\usepackage{microtype}
\usepackage{wrapfig}
\definecolor{color_blue}{HTML}{E7EFFA}
\definecolor{color_green}{HTML}{E6F8E0}
\definecolor{color_gray}{HTML}{ECECEC}
\definecolor{pearDark}{HTML}{2980B9}
\usepackage{graphicx}
\usepackage{amsmath}

\usepackage[utf8]{inputenc}
\usepackage{xcolor}
\usepackage{listings}
\usepackage[most]{tcolorbox}
\usepackage{caption}
\usepackage{algorithm}
\usepackage{algorithmic}
\usepackage{amsmath}
\usepackage{amssymb}
\usepackage{marvosym}

\lstdefinestyle{prompt}{
    basicstyle=\ttfamily\fontsize{7pt}{8pt}\selectfont,
    frame=none,
    breaklines=true,
    backgroundcolor=\color{lightgray},
    breakatwhitespace=true,
    breakindent=0pt,
    escapeinside={(*@}{@*)},
    numbers=none,
    numbersep=5pt,
    xleftmargin=5pt,
    aboveskip=2pt,
    belowskip=2pt,
}

\tcbset{
  aibox/.style={
    top=10pt,
    colback=white,
    enhanced,
    center,
  }
}
\newtcolorbox{AIbox}[2][]{aibox, title=#2,#1}
\definecolor{color_green}{HTML}{E6F8E0}
\title{Flow-OPD: On-Policy Distillation for Flow Matching Models}
\author{%
    Zhen Fang$^{1 *}$\,
    Wenxuan Huang$^{*\dag}$\textsuperscript{\Letter}\,
    Yu Zeng$^1$\,
    Yiming Zhao$^1$\,
    Shuang Chen$^2$\,
    \textbf{Kaituo Feng}$^3$  \\
    \textbf{Yunlong Lin}$^3$\,
    \textbf{Lin Chen}$^1$\,
    \textbf{Zehui Chen}$^1$\,
    \textbf{Shaosheng Cao}$^4$\textsuperscript{\Letter}\,
    \textbf{Feng Zhao}$^1$\textsuperscript{\Letter} \\
    {\normalsize$^1$University of Science and Technology of China} \quad
    {\normalsize$^2$University of California, Los Angeles} \\
    {\normalsize$^3$The Chinese University of Hong Kong} \quad 
    {\normalsize$^4$Xiaohongshu Inc.} \\
    {\tt fazii@mail.ustc.edu.cn (Zhen Fang)}  \quad
    {\tt wxhuang0616@gmail.com (Wenxuan Huang)} \\
    *: Equal Contribution\quad
    \dag: Project Leader\quad
    \Letter: Corresponding Author \\
    Project Page: \textcolor{blue}{\url{https://costaliya.github.io/Flow-OPD/}}
}

\begin{document}

\maketitle

\begin{abstract}
Existing Flow Matching (FM) text-to-image models suffer from two critical bottlenecks under multi-task alignment: the reward sparsity induced by scalar-valued rewards, and the gradient interference arising from jointly optimizing heterogeneous objectives, which together give rise to a ``seesaw effect" of competing metrics and pervasive reward hacking. Inspired by the success of On-Policy Distillation (OPD) in the large language model community, we propose \textbf{Flow-OPD}, the first unified post-training framework that integrates on-policy distillation into Flow Matching models. Flow-OPD adopts a two-stage alignment strategy: it first cultivates domain-specialized teacher models via single-reward GRPO fine-tuning, allowing each expert to reach its performance ceiling in isolation; it then establishes a robust initial policy through a Flow-based Cold-Start scheme and seamlessly consolidates heterogeneous expertise into a single student via a three-step orchestration of on-policy sampling, task-routing labeling, and dense trajectory-level supervision. We further introduce Manifold Anchor Regularization (MAR), which leverages a task-agnostic teacher to provide full-data supervision that anchors generation to a high-quality manifold, effectively mitigating the aesthetic degradation commonly observed in purely RL-driven alignment. Built upon Stable Diffusion 3.5 Medium, Flow-OPD raises the GenEval score from 63 to 92 and the OCR accuracy from 59 to 94, yielding an overall improvement of roughly 10 points over vanilla GRPO, while preserving image fidelity and human-preference alignment and exhibiting an emergent "teacher-surpassing" effect. These results establish Flow-OPD as a scalable alignment paradigm for building generalist text-to-image models.
The codes and weights will be released in: \url{https://github.com/CostaliyA/Flow-OPD}.
\end{abstract}

\section{Introudction}
Flow Matching (FM)~\cite{batifol2025flux,esser2024scaling,lipman2022flow,fang2025dualvla} has emerged as a superior paradigm for generative modeling, outperforming traditional diffusion models in both sampling efficiency and high-fidelity synthesis by learning continuous-time velocity fields.
However, as the research frontier shifts from unconstrained image synthesis toward highly-controllable, multi-dimensional alignment, the limitations of current post-training methodologies have become painfully evident. Modern applications demand that a single model masters a diverse spectrum of tasks—ranging from precise text rendering and complex compositional reasoning~\cite{huang2026visionr1incentivizingreasoningcapability, huang2026vision, chen2025advancing, chen2025ares, guo2025deepseek, chen2026opensearchvlopenrecipefrontier} to rigorous adherence to nuanced human aesthetic preferences—all within a unified generative space~\cite{han2026unicorn, chen2026unify,feng2026gen, huang2025interleaving}.

Recent advances have attempted to bridge this gap by porting Reinforcement Learning (RL) algorithms, such as Group Relative Policy Optimization (GRPO)~\cite{r1}, to the flow-matching domain~\cite{liu2025flow,xue2025dancegrpo,li2025mixgrpo}\footnote{In this paper, GRPO is used by default as Flow-GRPO in flow matching.}. These methods have demonstrated significant potential in single-reward scenarios, where on-policy exploration allows the model to refine its sampling trajectories and improve specific metrics like PickScore or aesthetic scores.
Nevertheless, different tasks demand heterogeneous and conflicting feature representations. As noted in LLM alignment~\cite{zeng2026glm}, sparse scalar rewards lack the granularity to harmonize these objectives, inducing a zero-sum "seesaw effect" where optimizing specific features (e.g., OCR) inevitably degrades aesthetics via reward hacking. This necessitates a shift to dense, trajectory-level distillation to provide uncoupled expert supervision.

This issue has recently found a compelling solution in the field of Large Language Models (LLMs):  On-Policy Distillation (OPD).  Benefiting from OPD, models such as DeepSeek-V4~\cite{guo2025deepseek}, Mimo v2~\cite{xiao2026mimo}, and GLM-5~\cite{zeng2026glm} successfully harmonize complex, multi-domain capabilities by distilling from specialized experts. This paradigm shift raises a pivotal question for the vision community: \textbf{\textit{Can Flow Matching models similarly leverage OPD to integrate the diverse strengths of multiple teacher models into a single, robust student model?}}
To address this pivotal question, we introduce Flow-OPD, the first framework to integrate OPD into the post-training pipeline of FM models. We propose a two-stage alignment strategy that begins by cultivating specialized domain teachers through single-reward GRPO fine-tuning, ensuring each expert reaches its performance ceiling in isolation. To facilitate a smooth transition for the student model, we develop a Flow-based Cold-Start strategy featuring two distinct variants—Supervised Fine-Tuning (SFT) based initialization and Model Merging—designed to establish a robust foundational policy capable of multi-task learning. Building upon this foundation, we apply OPD to the flow-matching process via a three-step orchestration: (1) performing \textbf{on-policy sampling} to capture the student model’s current velocity field, (2) executing task routing labeling where diverse experts provide dense supervision for respective domains, and (3) introducing Manifold Anchor Regularization (MAR), which incorporates a task-agnostic teacher to provide full-data supervision, effectively anchoring the generation process to a high-quality manifold and further elevating the aesthetic integrity of the synthesized images. 
Experimental results across multiple benchmarks and metrics demonstrate that Flow-OPD achieves 10\% improvement over vanilla GRPO with sparse rewards, establishing a new frontier for scaling alignment in flow-based generative models.
In summary, our contributions are three-fold:

\begin{figure*}[!t]
    \centering
    \includegraphics[scale=0.435]{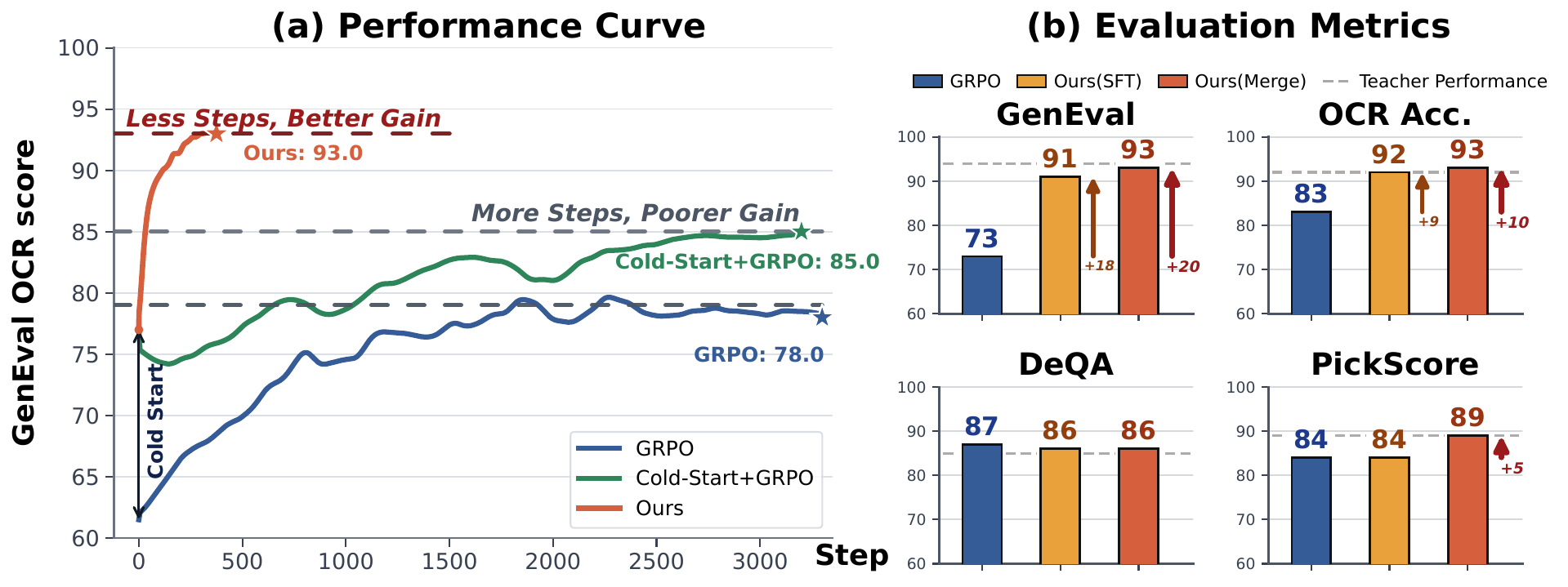}
    \caption{\textbf{Performance Comparison in Multi-task Training}. During training, Flow-OPD exhibits a steady increase in mean rewards across GenEval~\cite{geneval} and OCR~\cite{textdiffuser} benchmarks, reaching a peak of 93. In contrast, vanilla GRPO converges prematurely around 79. Our approach significantly outperforms GRPO in both image synthesis and text rendering while maintaining superior generation quality and human preference alignment. The curves are smoothed for visual clarity. DeQA and PickScore are norm to 0-1. We employ model merging for cold-start in the left subgraph. \textbf{The training steps are normalized to a 4-node NVIDIA H800 GPU configuration.}
    }
    \label{fig:mix}
\end{figure*}

\begin{itemize}
    \vspace{-0.2cm}
    \item \textbf{Analysis of Multi-task FM Training:} We provide a empirical  analysis of the failure modes of GRPO-based multi-task training in Flow Matching models, specifically identifying the challenges of reward sparsity and gradient interference. To resolve these, we are the first, to our best knowledge, to introduce OPD paradigm into the post-training of FM models.
    \item \textbf{The Flow-OPD Framework:} We propose \text{Flow-OPD}, a two-stage post-training framework that decouples expertise acquisition from model unification. Our framework introduces a Flow-based Cold-Start strategy (SFT and Merging variants), a task routing dense labeling mechanism for fine-grained supervision, and a novel Manifold Anchor Regularization (MAR) to ensure global generative quality through task-agnostic guidance.
    \item \textbf{Superior Performance and Generalization:} Through extensive experiments on four mainstream benchmarks, we demonstrate that Flow-OPD achieves a substantial 10-point improvement over the GRPO baseline. Notably, the unified student model matches or even surpasses the performance of specialized teachers in-domain, while exhibiting exceptional out-of-distribution (OOD) generalization capabilities.
\end{itemize}
\section{Related Work}
\paragraph{RL for T2I Models}
The success of RL-based alignment in large language models has recently inspired reinforcement learning for text-to-image (T2I) generation. Early methods such as DDPO~\cite{black2024training}, DPOK~\cite{fan2023dpok}, and ImageReward/ReFL~\cite{xu2023imagereward} formulate diffusion generation as policy optimization with rewards for aesthetics, human preference, or text-image alignment, while Diffusion-DPO~\cite{wallace2023diffusiondpo} aligns diffusion models using preference pairs. More recent GRPO-style methods extend RL to modern visual generators, including those for flow models~\cite{liu2025flowgrpo,xue2025dancegrpo}, and AR paradigms~\cite{yuan2025argrpo,zhang2025gcpo,ma2025stage,zhang2025maskfocus,ma2026margrpo}. However, T2I generation requires multiple rewards to cover aesthetics, alignment, fidelity, and compositional correctness. Existing solutions remain hard to control: DanceGRPO~\cite{xue2025dancegrpo} directly mixes rewards such as HPS and CLIP, often trading off one metric against another; Flow-GRPO~\cite{liu2025flowgrpo} uses staged reward/dataset curricula, making results sensitive to ordering and stage design; and GDPO~\cite{liu2026gdpo} shows that GRPO~\cite{guo2025deepseek} may suffer from reward-normalization collapse under multi-reward settings. This motivates a more controllable multi-reward coordination mechanism.

\paragraph{On-Policy Distillation}
Traditional offline distillation relies on fixed datasets and fails to adapt to the student's evolving trajectory. In contrast, On-Policy Distillation (OPD) dynamically couples the teacher's supervisory signal with the student’s exploration space. In the LLM domain, OPD has seen rapid development: GKD~\cite{agarwal2024policy} established the canonical framework to mitigate exposure bias; MiniLLM~\cite{gu2024minillm} and DistiLLM~\cite{ko2025distillm} introduced Reverse and Skewed Kullback-Leibler (KL) to refine mode-seeking and optimization stability; G-OPD~\cite{yang2026learning} unified OPD under KL-constrained RL theory; Entropy-Aware OPD~\cite{jin2026entropy} preserves diversity through adaptive divergence functions; Fast OPD~\cite{zhang2026fast} significantly accelerates computation via prefix truncation; and PACED~\cite{xu2026paced} implements a competence-aware curriculum based on gradient signal-to-noise analysis. Despite these LLM advancements, OPD remains underexplored in visual Flow Matching models, which require dense supervision within high-dimensional velocity fields. We propose Flow-OPD, the first systematic migration of on-policy distillation to Flow Matching, utilizing multi-teacher dense supervision to overcome the reward sparsity bottleneck.
\section{Preliminaries}
\paragraph{Flow-Matching Models}
Flow Matching (FM) maps a noise distribution $p_0$ to data $p_{\text{data}}$ via an ODE $\text{d}\mathbf{x}_t = v_t(\mathbf{x}_t, t) \text{d}t$. Under the Optimal Transport (OT) formulation~\cite{lipman2022flow}, the path is $\mathbf{x}_t = (1-t)\mathbf{x}_0 + t\mathbf{x}_1$, and the model $v_\theta$ learns the constant velocity $(\mathbf{x}_1 - \mathbf{x}_0)$ via:
\begin{equation}
    \mathcal{L}_{\text{FM}}(\theta) = \mathbb{E}_{t, \mathbf{x}_0, \mathbf{x}_1} \left[ \| v_\theta(\mathbf{x}_t, t) - (\mathbf{x}_1 - \mathbf{x}_0) \|^2 \right]
\end{equation}
Following Flow-GRPO~\cite{liu2025flow}, we conceptualize the discretized ODE integration as a sequential \textit{Markovian denoising process}. By formulating each transition $\mathbf{x}_{t} \to \mathbf{x}_{t+\Delta t}$ as a Markovian state step, this perspective bridges continuous generative dynamics with reinforcement learning, defining a formal trajectory for step-wise policy optimization.

\paragraph{On-Policy Distillation}
Knowledge distillation aims to compress teacher capabilities into a student model by minimizing their output divergence. To mitigate distribution shift, on-policy distillation (OPD)~\cite{lu2025onpolicydistillation} requires the student $f_\theta$ to generate trajectories $\tau \sim p_\theta(\tau)$ under the guidance of real-time teacher supervision. For Autoregressive (AR) models, this optimization is formulated as minimizing the Reverse Kullback-Leibler (KL) divergence between the student and teacher distributions:
\begin{equation}
\label{opd}
    \mathcal{L}_{\text{OPD}} = -\mathbb{E}_{y \sim \pi_\theta} \left[ \log \frac{\pi_{\text{teacher}}(y|x)}{\pi_{\theta}(y|x)} \right] = D_{\text{KL}}(\pi_{\theta} \| \pi_{\text{teacher}})
\end{equation}
By aligning the model on its own generated distribution, OPD effectively suppresses exposure bias and ensures robust generalization in interactive or iterative generation tasks.
\section{Motivation}

\subsection{Question 1: Why GRPO Works?}
Standard FM relies on offline reconstruction, fundamentally limiting performance to static dataset quality and failing to optimize non-differentiable preferences. GRPO~\cite{guo2025deepseek,liu2025flow,xue2025dancegrpo} overcomes this via \textbf{online exploration}. By actively sampling $G$ outputs from its current policy $\pi_\theta$, it evaluates self-generated states using a Group Relative Advantage, $A(\mathbf{x}_1^{(i)}) = (r(\mathbf{x}_1^{(i)}) - \mu)/\sigma$. The policy gradient is then explicitly driven by these online experiences:
\begin{equation}
    \nabla_\theta J(\theta) \approx \frac{1}{G} \sum_{i=1}^G A(\mathbf{x}_1^{(i)}) \nabla_\theta \log p_\theta(\mathbf{x}_1^{(i)} | c)
\end{equation}
This continuous exploration of its own dynamic distribution enables the model to discover novel, high-reward trajectories, successfully breaking the performance ceiling of offline SFT.

\subsection{Question 2: Why GRPO Failed? A Multi-Task Perspective}
\begin{figure*}[t]
    \centering
    \includegraphics[scale=0.34]{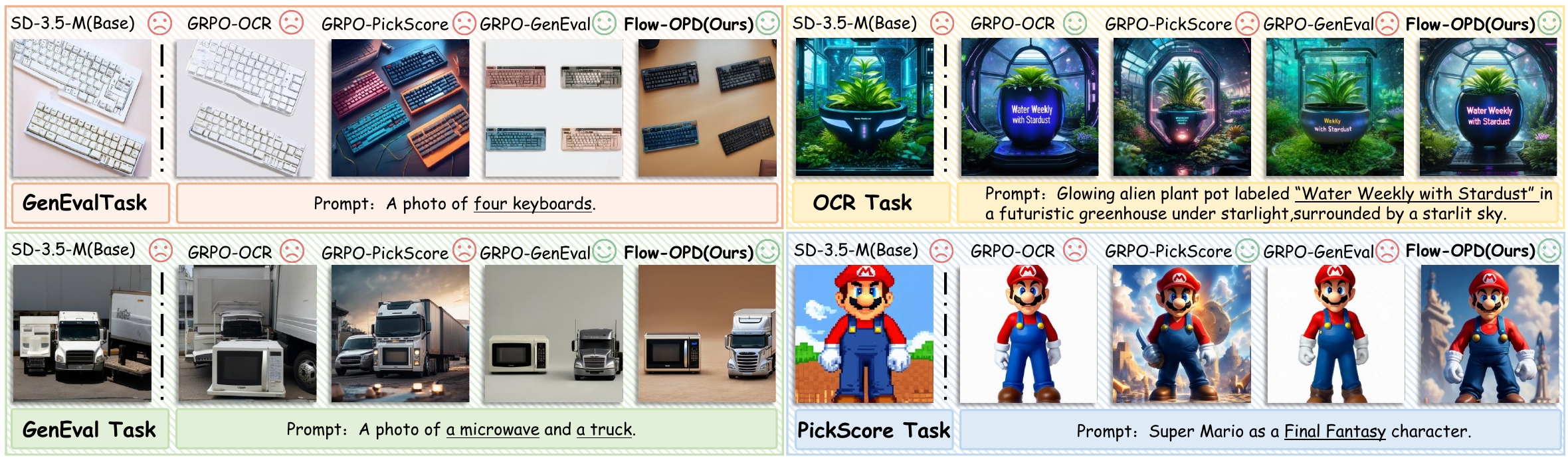}
    \caption{
Cross-task evaluation of single-reward GRPO. Optimizing with a solitary reward signal severely compromises generalization, leading to capability degradation on non-target metrics. All baseline setups strictly adhere to the official Flow-GRPO implementation.} 
    \label{fig:intro}
\end{figure*}
Despite its target-specific efficacy, single-reward GRPO incurs severe \textbf{degradation in orthogonal capabilities} (Fig.~\ref{fig:intro}). This catastrophic forgetting stems from \textbf{unconstrained gradient interference} driven by sparse scalar rewards within a shared parameter space $\theta$.

For a parameter update $\Delta \theta$ driven by a target task $\mathcal{T}_1$ with advantage $A_1$, the collateral impact on an unmonitored capability $\mathcal{T}_k$ ($k \neq 1$) can be approximated via first-order Taylor expansion:
\begin{equation}
\Delta \mathcal{J}_k \approx \langle \nabla_\theta \mathcal{J}_k, \Delta \theta \rangle \propto \mathbb{E}_{\mathbf{x} \sim \pi_\theta} \left[ A_1(\mathbf{x}) \left\langle \nabla_\theta \mathcal{J}_k, \nabla_\theta \log \pi_\theta(\mathbf{x} | c) \right\rangle \right]
\end{equation}

In high-dimensional spaces, divergent task gradients frequently conflict ($\langle \nabla_\theta \mathcal{J}_k, \nabla_\theta \mathcal{J}_1 \rangle < 0$). Lacking supervisory signals for $\mathcal{T}_k$, the optimizer aggressively exploits these unmonitored degrees of freedom to maximize $A_1$, dismantling pre-trained synergies and leading to manifold collapse. This prompts a natural question: \textit{Can we resolve this degradation by simply mixing multiple datasets and rewards for joint optimization?}

\subsection{Question 3: Can mix training solve the problem?}

\noindent 
\begin{minipage}[t]{0.48\textwidth} 
    To explore the feasibility of mix training approach, we conduct a controlled empirical experiment on Stable Diffusion 3.5 Medium (SD-3.5-M)~\cite{esser2024scaling}. Following Flow-GRPO, we progressively stack four distinct reward functions: GenEval, OCR, PickScore, and DeQA.    As demonstrated in Table~\ref{tab:toy_intro}, mixing scalar rewards fails to construct a stable cognitive foundation.

\end{minipage}
\hfill 
\begin{minipage}[t]{0.48\textwidth} 
    \vspace{-12pt} 
    \centering
    \makeatletter\def\@captype{table}\makeatother 
    \caption{Capability degradation in multi-reward optimization.}
    \renewcommand{\arraystretch}{0.9} 
    \label{tab:toy_intro}
    \small
    \begin{tabular}{lll}
        \toprule
        \textbf{Model} & \textbf{GenEval} & \textbf{OCR} \\
        \midrule
        SD-3.5-M    & 0.63 & 0.59 \\
        +GenEval    & 0.94 & 0.65 \\
        +OCR        & 0.89 \drop{5} & 0.91 \\
        +PickScore  & 0.82 \drop{7} & 0.86 \drop{5} \\
        +DeQA       & 0.73 \drop{9} & 0.83 \drop{3} \\
        \bottomrule
    \end{tabular}
\end{minipage}

While the initial reward (+GenEval) succeeds, subsequent additions trigger catastrophic forgetting (e.g., +OCR degrades GenEval by 5\%). This corroborates our hypothesis of \textbf{Gradient Interference} ($\langle \nabla_\theta \mathcal{J}_i, \nabla_\theta \mathcal{J}_j \rangle < 0$). Compressing multi-dimensional conflicts into a scalar advantage forces a zero-sum game; for instance, accommodating aesthetic stylization (PickScore) aggressively overwrites precise geometric representations.
Consequently, scalar reward mixing is fundamentally unscalable due to this sparse \textbf{Information Bottleneck}. To avoid parameter cannibalization, we require a supervisory signal that is simultaneously \textbf{on-policy} (maintaining exploration) and \textbf{densely uncoupled} (preventing interference). Inspired by Multi-Teacher On-Policy Distillation (OPD) in LLMs, we propose \textbf{Flow-OPD}. This framework seamlessly introduces the multi-teacher paradigm into continuous Foundation Models, achieving active on-policy exploration guided by dense supervision.

\section{Method: Flow-OPD}
Flow-OPD reformulates multi-task alignment via dense supervision on self-generated trajectories. We first train domain-expert teachers using Flow-GRPO. Following cold-start initialization, the student undergoes Multi-Teacher Online Distillation, dynamically routing online samples to specific teachers for fine-grained guidance. Finally, Manifold Anchor Regularization decouples functional alignment from aesthetic collapse, preserving the inherent generative prior.

\subsection{Cold Start}
To ensure a stable initialization $\theta_0$ and prevent trajectory divergence during early rollout, we explore two cold-start strategies: SFT-based and model-merging initialization. Our SFT protocol follows Flow-GRPO but utilizes trajectories sampled from specialized teachers, ensuring the student inherits expert-level knowledge distributions from the outset. Alternatively, model merging superposes the anisotropic priors of divergent teachers into a unified parameter state. This "merging-as-initialization" approach positions the student in a high-competence region of the loss landscape, where multi-task synergies are already nascent, providing a robust foundation for subsequent distillation.

\subsubsection{Multi-Teacher On Policy Distillation}

\paragraph{Bridging OPD and Flow Matching}
As shown in Eq.~\ref{opd}, ThinkingMachines' OPD~\cite{lu2025onpolicydistillation} optimizes a student policy $\pi_\theta$ by utilizing the Reverse KL divergence against a teacher distribution $\pi_{\phi}$ as an environment reward over autonomously generated trajectories $\tau$.
To transpose this Policy Gradient (PG) paradigm into the continuous-time FM framework, we map the discrete token sequence to the continuous latent trajectory $x_t \in \mathbb{R}^d$. The ar prediction translates to the instantaneous transition policy parameterized by the velocity field $v_\theta(x_t, t)$. Crucially, instead of directly minimizing the distance between vector fields via supervised regression, we derive the exact continuous-time KL divergence and utilize it as a \textit{dense reward signal} to guide policy exploration via PG.

\paragraph{On Policy Sampling}
The fundamental premise of Flow-OPD requires the student to expose its own specific distribution shifts. To facilitate sufficient state-space exploration—a necessity for escaping local optima in RL—we inject stochasticity by converting the deterministic probability flow ODE into an equivalent Stochastic Differential Equation (SDE)~\cite{liu2025flow}:
\begin{equation}
\text{d}x_t = \left[ v_{\theta}(x_t, t) + \frac{\sigma_t^2}{2t}(x_t + (1-t)v_{\theta}(x_t, t)) \right] \text{d}t + \sigma_t \text{d}w
\end{equation}
Applying Euler-Maruyama discretization over a time step $\Delta t$, the student's transition behavior acts as a local isotropic Gaussian policy:
\begin{equation}
\pi_\theta(x_{t-\Delta t} | x_t, c) = \mathcal{N}(\mu_\theta(x_t, t), \sigma_t^2 \Delta t I)
\end{equation}
By sampling $G$ independent trajectories per prompt, this generates an on-policy marginal distribution $x_t \sim \rho_t^\theta(\cdot | c)$, acting as the stochastic behavioral policy.

\paragraph{Task-Specific Teacher Labeling}
At each explored state $x_t$, the student queries the ensemble of expert teachers for localized supervision. To eliminate inter-domain gradient interference, we implement a \textbf{hard routing mechanism} $\mathbb{1}_{\mathcal{T}(c)=k}$, which maps the textual condition $c$ to its unique corresponding domain expert $k$ among the ensemble. This mechanism selectively activates a single teacher to provide the reference velocity field $v_{\phi_k}(x_t, t, c)$. The target flow is thus defined as:
\begin{equation}
v_{\text{target}}(x_t, t, c) = v_{\phi_k}(x_t, t, c), \quad \text{where } k = \mathcal{R}(c)
\end{equation}
where $\mathcal{R}(\cdot)$ denotes the deterministic task-to-teacher routing function. This yields a task-specific target transition policy $\pi_{\text{target}} = \mathcal{N}(\mu_{\text{target}}(x_t, t), \sigma_t^2 \Delta t I)$ that serves as the definitive gold standard for evaluating the student's on-policy trajectories.

\paragraph{Deriving the Analytical KL in Flow Models}
In LLMs, OPD defines the dense advantage as the negative Reverse KL divergence: $A_t = -D_{\text{KL}}(\pi_\theta \| \pi_{\text{target}})$. Because the discrete token vocabulary prohibits gradients from flowing directly through the sampling process, the optimization strictly relies on score-function estimators (e.g., the Policy Gradient theorem):
\begin{equation}
\nabla_\theta \mathcal{J}_{\text{LLM}} = \mathbb{E}_{s_t \sim \pi_\theta} \Big[ \nabla_\theta \log \pi_\theta(a_t|s_t) \cdot A_t \Big]
\end{equation}
This discrete formulation inevitably introduces high gradient variance, requiring computationally expensive \texttt{logprob} tracking, importance sampling, and complex surrogate clipping mechanisms (e.g., PPO) to stabilize training.
To migrate this framework to flow matching models, we must mathematically adapt the KL computation to continuous state spaces. Following the SDE formulation, both the student transition policy $\pi_\theta(\cdot|x_t)$ and the target policy $\pi_{\text{target}}(\cdot|x_t)$ are $d$-dimensional multivariate Gaussian distributions. Crucially, they share the exact same isotropic covariance matrix $\Sigma_t = \sigma_t^2 \Delta t I$ induced strictly by the forward SDE noise schedule. 
For two general Gaussian distributions $\mathcal{N}_1(\mu_1, \Sigma_1)$ and $\mathcal{N}_2(\mu_2, \Sigma_2)$, the expected KL divergence is analytically defined as:
\begin{equation}
D_{\text{KL}}(\mathcal{N}_1 \| \mathcal{N}_2) = \frac{1}{2} \left[ \text{tr}(\Sigma_2^{-1} \Sigma_1) - d + (\mu_1 - \mu_2)^T \Sigma_2^{-1} (\mu_1 - \mu_2) + \ln\left(\frac{\det \Sigma_2}{\det \Sigma_1}\right) \right]
\end{equation}
Because our policies share identical covariances ($\Sigma_1 = \Sigma_2 = \Sigma_t$), the trace term evaluates to the data dimension $\text{tr}(I) = d$, which seamlessly cancels out the $-d$ constant. Simultaneously, the logarithmic ratio of identical determinants evaluates to $\ln(1) = 0$. Consequently, the complex divergence collapses exclusively into the Mahalanobis distance. Substituting our specific inverse covariance $\Sigma_t^{-1} = (\sigma_t^2 \Delta t)^{-1} I$, the KL divergence gracefully reduces to the $L_2$ distance between their means~\cite{liu2025flow}:
\begin{equation}
D_{\text{KL}}(\pi_\theta \| \pi_{\text{target}}) = \frac{\| \mu_\theta(x_t, t) - \mu_{\text{target}}(x_t, t) \|^2}{2\sigma_t^2 \Delta t}
\end{equation}

Following the Euler-Maruyama discretization, the policy mean $\mu_\theta(x_t, t)$ is deterministically parameterized by the predicted vector field $v_\theta$:
\begin{equation}
\mu_\theta(x_t, t) = x_t + \left[ v_\theta(x_t, t, c) + \frac{\sigma_t^2}{2t} \big(x_t + (1-t)v_\theta(x_t, t, c)\big) \right] \Delta t
\end{equation}
When subtracting the target mean $\mu_{\text{target}}$, the state-dependent base terms identically cancel out. The KL divergence elegantly simplifies into a scaled discrepancy strictly between the vector fields:
\begin{equation}
D_{\text{KL}}(\pi_\theta \| \pi_{\text{target}}) = \frac{\Delta t}{2} \left( \frac{\sigma_t(1-t)}{2t} + \frac{1}{\sigma_t} \right)^2 \| v_\theta(x_t, t, c) - v_{\text{target}}(x_t, t, c) \|^2
\end{equation}

\paragraph{Bringing OPD to Flow-Matching Models}
In LLMs, because the discrete token vocabulary prohibits direct differentiation through the sampling process, OPD must rely on score-function estimators (i.e., the Policy Gradient theorem) to optimize the dense reward $r_t = -D_{\text{KL}}(\pi_\theta \| \pi_{\text{target}})$:
\begin{equation}
\nabla_\theta \mathcal{J}_{\text{LLM-PG}} = \mathbb{E}_{s_t \sim \pi_\theta} \Big[ \nabla_\theta \log \pi_\theta(a_t|s_t) \cdot \big( -D_{\text{KL}}(\pi_\theta \| \pi_{\text{target}}) \big) \Big]
\label{equ:llm_pg}
\end{equation}
Directly migrating this policy gradient formulation to the continuous state space of flow models yields a structurally identical Flow-OPD PG update:
\begin{equation}
\nabla_\theta \mathcal{J}_{\text{Flow-OPD}} = \mathbb{E}_{x_t \sim \text{SDE}_\theta} \Big[ \nabla_\theta \log \pi_\theta(x_{t-\Delta t} | x_t) \cdot \big( -D_{\text{KL}}(\pi_\theta \| \pi_{\text{target}}) \big) \Big]
\label{equ:pg}
\end{equation}

In discrete spaces, evaluating Eq.~\ref{equ:llm_pg} necessitates high-variance Monte Carlo estimations. However, a fundamental property of the Log-Derivative trick dictates that the expected policy gradient of the KL divergence mathematically equals the direct gradient of the expected KL divergence itself. Specifically, computing the PG expectation over an infinite number of samples mathematically converges to exactly: $\mathbb{E}[\nabla_\theta \log \pi_\theta \cdot (-D_{\text{KL}})] \equiv \nabla_\theta (-D_{\text{KL}})$.

While this direct gradient is analytically intractable for discrete LLMs, our continuous flow formulation possesses a fully differentiable, closed-form solution. Consequently, we can entirely bypass the high-variance PG estimator, the $\log \pi_\theta$ computations, and PPO surrogate bounds. \textbf{Fundamentally, directly minimizing this exact MSE loss in flow models is mathematically strictly equivalent to executing the Policy Gradient OPD in LLMs, but effectively reduces the gradient variance to zero.} Thus, Eq.~\ref{equ:pg} can be simplified by directly backpropagating through the standard time-weighted MSE loss:
\begin{equation}
\nabla_\theta \mathcal{L}_{\text{Flow-OPD}} = \nabla_\theta \mathbb{E}_{x_t \sim \text{SDE}_\theta} \left[ w(t) \| v_\theta(x_t, t, c) - \text{SG}\big(v_{\text{target}}(x_t, t, c)\big) \|^2 \right]
\end{equation}
where $v_\theta$ remains attached to the computation graph to receive gradients, $\text{SG}(\cdot)$ denotes the stop-gradient operator for the teacher's target, and $w(t) = \frac{\Delta t}{2} \big( \frac{\sigma_t(1-t)}{2t} + \frac{1}{\sigma_t} \big)^2$. Ultimately, this formulation seamlessly unites the rigorous state-space exploration of on-policy RL with the deterministic optimization efficiency of regression.

\paragraph{Manifold Anchor Regularization}
Aggressively optimizing for functional targets (e.g., precise text rendering or strict spatial layout) frequently induces reward hacking, manifesting as a severe degradation in visual aesthetics and generative diversity~\cite{liu2025flow}. To decouple functional alignment from stylistic collapse, we introduce a continuous-time aesthetic preservation mechanism inspired by the KL penalty in Flow-GRPO. 

However, rather than anchoring to a generic pre-trained model, we maintain a frozen \textit{aesthetic teacher} (e.g., optimized via DeQA) to provide a high-fidelity regularizing vector field $v_{\text{base}}$. As previously derived, the Reverse KL divergence in the SDE framework elegantly translates to the time-weighted $L_2$ distance between vector fields. In our implementation, the optimization is formulated as minimizing a total loss $\mathcal{L}_{\text{Total}}(\theta)$, which is the direct sum of the policy loss $\mathcal{L}_{\text{Policy}}(\theta)$ (defined as the negative of the surrogate objective $-\mathcal{J}(\theta)$) and this dense KL penalty:
\begin{equation}
\mathcal{L}_{\text{Total}}(\theta) = \mathcal{L}_{\text{Flow-OPD}}(\theta) + \lambda \mathbb{E}_{c, t, x_t \sim \rho_t^\theta} \left[ w(t) \| v_\theta(x_t, t, c) - v_{\text{aesthetic}}(x_t, t, c) \|^2 \right]
\end{equation}
This KL regularization operates as a continuous elastic anchor. It guarantees that while the student policy greedily absorbs the functional intelligence from the multi-teacher ensemble, it remains strictly bounded to a high-quality visual manifold, completely averting the aesthetic degradation typical in single-objective RL.
\section{Experiments}
\subsection{Experimental Setup}
Following Flow-GRPO~\cite{liu2025flow}, we evaluate our method on four tasks: GenEval~\cite{geneval}, OCR~\cite{textdiffuser}, PickScore~\cite{pickscore}, and DeQA~\cite{deqa}. We adopt the official checkpoints as expert teachers for the first three tasks. The DeQA teacher is specifically trained across the three datasets by blending DeQA and PickScore rewards at a 4:6 ratio. All training and test data strictly follow the Flow-GRPO splits. Training is executed on 4 nodes ($8 \times \text{H800}$ GPUs each), while evaluation is conducted on a single $8 \times \text{H800}$ node.

We primarily evaluate Flow-OPD against two categories of baselines: 
(1) \textbf{Monolithic-Reward GRPO}, denoted as \textit{GRPO-[reward name]}, where the model is fine-tuned using Flow-GRPO on a single reward objective; 
(2) \textbf{Hybrid-Reward GRPO}, denoted as \textit{GRPO-Mix}, which employs a weighted reward combination with a fixed ratio of \textit{GenEval} : \textit{OCR} : \textit{PickScore} = 3 : 1 : 1. 
These baselines serve to highlight the limitations of conventional scalar-based alignment when scaling to multi-dimensional expert capabilities.
\subsection{Main Results}

\begin{figure*}[t]
    \centering
    \includegraphics[scale=0.49]{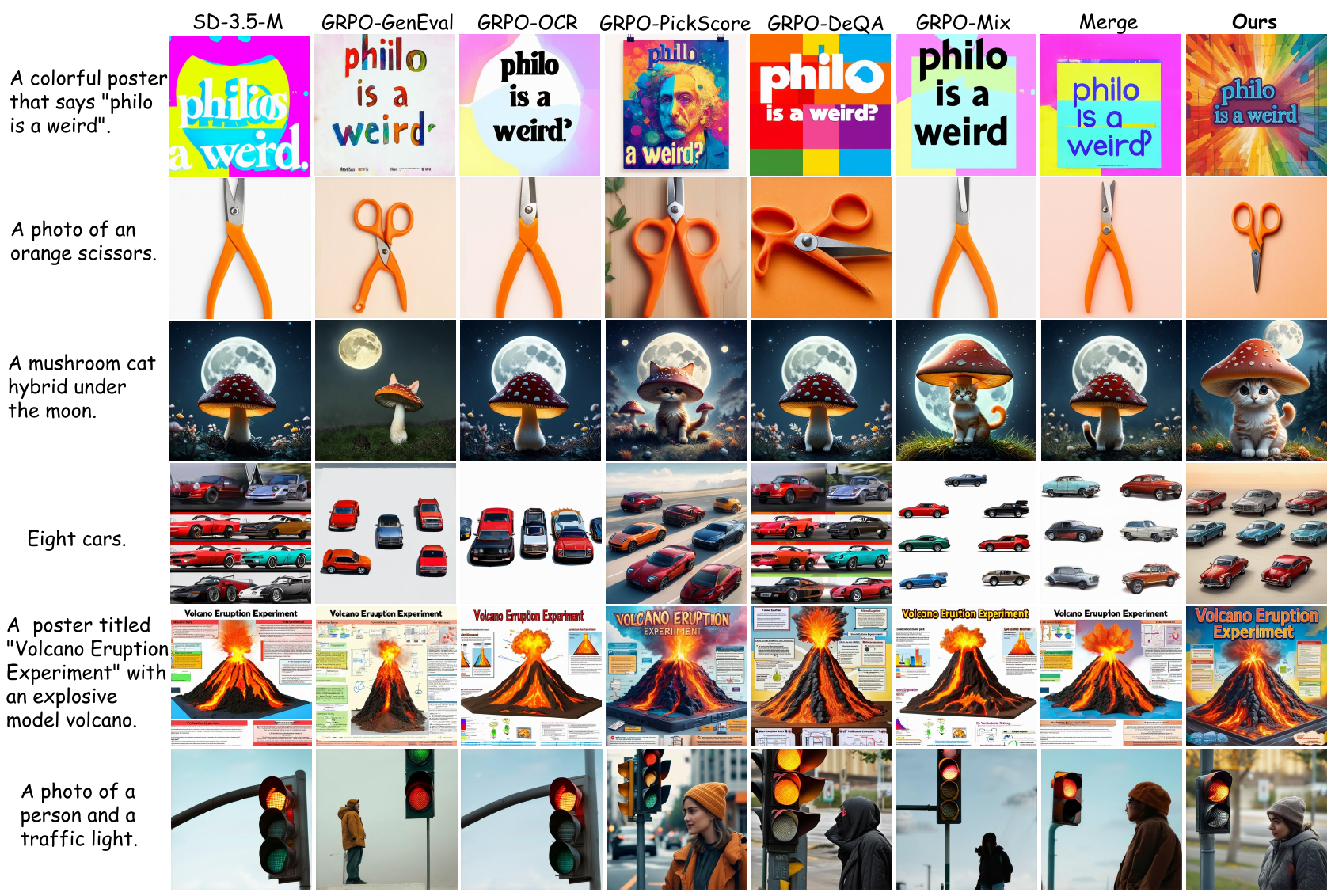}
\caption{Qualitative comparison between Flow-OPD and various baselines across diverse tasks. Our method consistently demonstrates superior instruction-following capabilities, delivering high-fidelity image synthesis and structural coherence that align more closely with human preferences.}
    \label{fig:main}
\end{figure*}

\begin{table}[t]
    \centering
    \caption{Model Performance Comparison on Compositional Image Generation, Visual Text Rendering, and Image Quality benchmarks. The avg values are computed by averaging four 0-1 normalized metrics. Scores of teacher models are \textbf{\underline{bolded and underlined}} to denote the performance ceiling and are excluded from the comparative. The best score is in \colorbox{color_blue}{blue} and the second best score is in \colorbox{color_green}{green}.}
    \label{tab:model_comparison_updated}
    \begin{tabular}{l|cccc|c}
        \toprule
        \textbf{Model} & \textbf{GenEval} & \textbf{OCR Acc.} & \textbf{DEQA} & \textbf{PickScore} & \textbf{Avg} \\
        \midrule
        SD-3.5-M        & 0.63            & 0.59            & 4.07          & 21.64               & 0.7165 \\
        \midrule
        +GRPO-Geneval   & \textbf{\underline{0.94}} & 0.65          & 4.01          & 21.53               & 0.8050 \\
        +GRPO-OCR       & 0.64            & \textbf{\underline{0.92}} & 4.06          & 21.69               & 0.8015 \\
        +GRPO-deqa      & 0.64            & 0.66            & \textbf{\underline{4.23}}         & \colorbox{color_green}{23.02}               & 0.7578 \\
        +GRPO-Pickscore & 0.51            & 0.69            & 4.22          & \textbf{\underline{23.19}} & 0.7340 \\             
        \midrule
        GRPO-Mix        & 0.73            & 0.83            & \colorbox{color_green}{4.33} & 21.84               & 0.8165 \\
        SFT+GRPO-Mix    & 0.85            & 0.86           & 4.29          & 21.79               & 0.8515 \\
        Merge+GRPO-Mix  & 0.84            & 0.86           & 4.18          & 21.87               & 0.8442 \\
        Ours (SFT)       & \colorbox{color_green}{0.91}           & \colorbox{color_green}{0.92} & 4.29          & 21.83               & \colorbox{color_green}{0.8820} \\ 
        Ours (Merge)     & \colorbox{color_blue}{0.93} & \colorbox{color_blue}{0.93} & \colorbox{color_blue}{4.31} & \colorbox{color_blue}{23.05} & \colorbox{color_blue}{0.9021} \\ 
        \bottomrule
    \end{tabular}
\end{table}

The quantitative results in Table~\ref{tab:model_comparison_updated} demonstrate that Flow-OPD consistently matches or surpasses the specialized teacher models across all benchmarks, particularly in text rendering and DeQA image quality. Crucially, it resolves the severe cross-domain interference inherent to specialization (e.g., the PickScore teacher's GenEval score dropping to 0.51) and overcomes the optimization bottlenecks of sparse-reward multi-task GRPO. By leveraging dense multi-expert supervision, Flow-OPD seamlessly consolidates diverse expertise without capability degradation.

Qualitative results in Fig.~\ref{fig:main}) show that Flow-OPD achieves an optimal multi-task trade-off, balancing high prompt fidelity with superior visual aesthetics. Remarkably, Flow-OPD succeeds in certain edge cases where all individual teachers fail, a phenomenon we term \textit{Teacher-Surpassing}. We hypothesize this emergent superiority stems from knowledge cross-pollination within the latent flow manifold. While individual teachers are constrained by domain-specific biases, simultaneous dense guidance forces the student to learn a more holistic, smoothed representation. This collective supervision bridges epistemic gaps, enabling the student to synthesize novel trajectories that ultimately surpass any single supervisor.
\subsection{Analysis}
\subsubsection{Cold Start Ablation}

\begin{figure*}[t]
    \centering
    \includegraphics[scale=0.58]{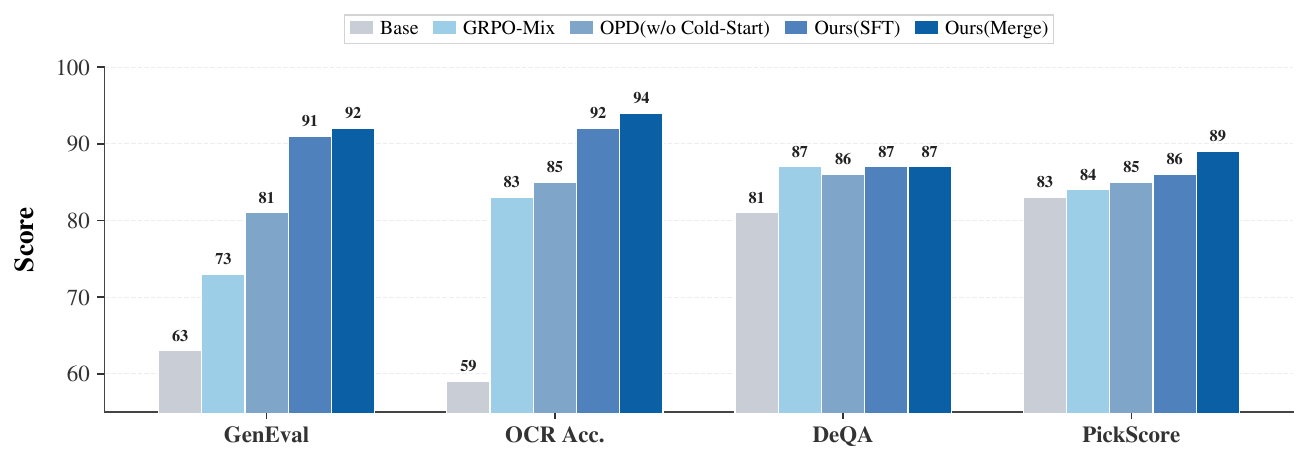}
    \caption{Cold-start ablation results.
    }
    \label{fig:cold—start}
\end{figure*}

\begin{figure*}[t]
    \centering
    \includegraphics[scale=0.445]{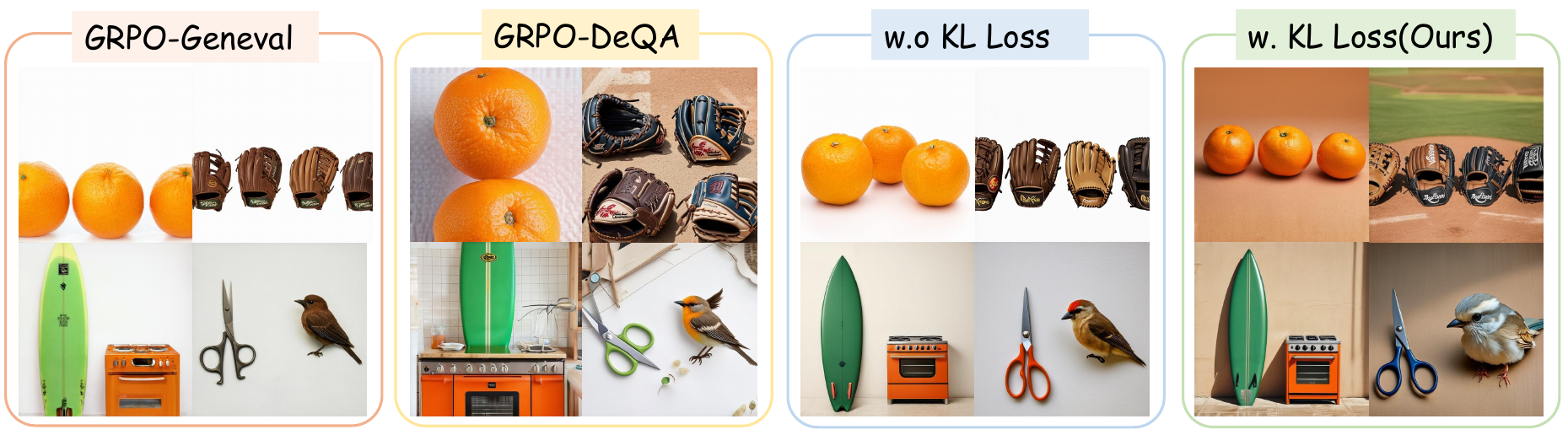}
    \caption{Qualitative ablation results of Manifold Anchor Regularization.
    }
    \label{fig: kl}
\end{figure*}

\begin{table}[t]
\centering
\small
    \caption{{\bf T2I-CompBench++ Result.} 
The best score is in \colorbox{pearDark!20}{blue}.}
\resizebox{\linewidth}{!}{
\begin{tabular}{lccccccc}
\toprule
\textbf{Model} & \textbf{Color} & \textbf{Shape} & \textbf{Texture} & \textbf{Complex} & \textbf{3D-Spatial} & \textbf{Numeracy} & \textbf{Non-Spatial} \\ \midrule
SD3.5-M~\cite{esser2024scaling}           & 0.7994 & 0.5669 & 0.7338 & 0.3800  & 0.3739 & 0.5927 & 0.3146 \\
GRPO-mix& 0.7966 & 0.5803 & 0.7392 & 0.3677  & 0.3681 & 0.6388 & 0.3130 \\
Cold Start      & 0.8173 & 0.6126 &0.7342 & 0.3870  & 0.4249 & 0.6458 & 0.3145 \\
Cold Start+GRPO         &  0.8031 & 0.5985 & 0.7409 & 0.3842  & 0.4017 & 0.6269 & 0.3136 \\
\textbf{Ours (Merge)} & \colorbox{pearDark!20}{0.8298} & \colorbox{pearDark!20}{0.6292} & \colorbox{pearDark!20}{0.7446} & \colorbox{pearDark!20}{0.3943} & \colorbox{pearDark!20}{0.4565} & \colorbox{pearDark!20}{0.6837} & \colorbox{pearDark!20}{0.3163} \\ \bottomrule
\end{tabular}
}
\vspace{-2mm}
\label{tab:t2i_comp_bench}
\end{table}

\begin{table}[!h]
\small
    \centering
    \caption{Performance Comparison on General Image Quality and Alignment Metrics. The best score is in \colorbox{color_blue}{blue}.}
    \label{tab:reward_metrics}
    \renewcommand{\arraystretch}{1} 
    \begin{tabular}{lccccc} 
        \toprule
        \textbf{Model} & \textbf{ImageReward} & \textbf{Aesthetic} & \textbf{UnifiedReward} & \textbf{HPS-v2.1} & \textbf{QwenVL Score}\\
        \midrule
        SD-3.5-M       & 1.02 & 5.87 & 3.339 & 0.2982&3.45 \\
        GRPO-DeQA      & 1.33 & 5.97 & 3.456 &0.2846 &3.68 \\
        GRPO-mix  & 1.23 & 5.93 & 3.501 & 0.3101&3.88 \\

        \midrule 
        w.o. MAR&1.26&5.89&3.518&0.2998&3.82\\
        \textbf{Ours (Merge)} & \colorbox{color_blue}{1.36} & \colorbox{color_blue}{6.23} & \colorbox{color_blue}{3.659} &\colorbox{color_blue}{0.3302}     &\colorbox{color_blue}{4.05} \\
        \bottomrule
    \end{tabular}
\end{table}

As shown in Fig.~\ref{fig:cold—start}, cold-start initialization rapidly establishes a robust foundation for subsequent training. Between the two regimes, SFT serves as a widely adopted and highly scalable strategy; notably, its inherent flexibility presents a promising avenue for extracting capabilities from heterogeneous teachers in future applications. Conversely, model merging optimally leverages the available homogeneous teachers for superior functional alignment without any additional training costs. 
Crucially, Flow-OPD consistently outperforms both from-scratch and cold-started multi-task GRPO. While GRPO converges to sub-optimal states due to inter-task conflicts caused by sparse scalar rewards, Flow-OPD leverages dense multi-expert supervision to resolve gradient interference. Consequently, our method achieves substantial, uniform gains across all baselines, successfully matching or exceeding the performance ceilings of individual specialized teachers.

\subsubsection{OOM Generalization}

To further investigate the generalization capabilities of our method, we conduct additional evaluations on the T2I-CompBench benchmark. Flow-OPD demonstrates superior out-of-domain generalization compared to multi-task GRPO, achieving state-of-the-art (SOTA) performance across multiple compositional metrics. Notably, when initialized from the identical cold-start baseline, standard GRPO suffers from catastrophic forgetting in specific capability dimensions, such as shape rendering and 3D spatial relations. In contrast, by leveraging dense multi-expert supervision and task-style decoupling regularization, Flow-OPD effectively mitigates these regression issues, yielding robust and comprehensive performance enhancements.

\subsubsection{Manifold Anchor Regularization}

Manifold Anchor Regularization (MAR) is a task-agnostic constraint designed to maintain generative integrity and aesthetic alignment. As shown in Fig.~\ref{fig: kl}, vanilla GRPO-based optimization often triggers background mode collapse—where models overfit to monotonous environments—and semantic redundancy, leading to identical features across multiple entities due to coarse reward granularity. While teachers like DeQA provide diverse samples, they often struggle with instruction following. MAR resolves these issues by anchoring optimization to a high-fidelity manifold, balancing structural diversity with precise semantic adherence.
Table~\ref{tab:reward_metrics} further provides quantitative evidence of our method's superiority in image quality and human preference alignment. The integration of MAR leverages additional supervision across the entire dataset, significantly enhancing both the visual quality and the expressive power of the generated images.
\section{Conclusion}

We introduced \textbf{Flow-OPD}, the first framework to integrate on-policy distillation into Flow Matching models, effectively resolving reward sparsity and gradient interference. By replacing scalar rewards with dense, trajectory-level supervision, Flow-OPD breaks the "seesaw effect" of competing metrics.
Our results on SD-3.5-M show that Flow-OPD successfully consolidates expertise in composition and typography while achieving an emergent "teacher-surpassing" effect. Through Manifold Anchor Regularization (MAR), the framework maintains high visual fidelity by decoupling functional alignment from aesthetic preservation. Ultimately, Flow-OPD provides a scalable paradigm for developing generalist text-to-image models with superior generative integrity.

\bibliographystyle{unsrt}
\bibliography{custom}
\newpage
\clearpage
\appendix
\section{More Details}
\label{Training}
Following the data and reward configurations of Flow-GRPO, we conducted multi-task hybrid training for GRPO-mix using an epoch ratio of 3:1:1 for GenEval, OCR, and PickScore, respectively. During each epoch, rewards were exclusively provided by the reward model corresponding to the current data partition. Training was executed on a distributed cluster of four nodes, each equipped with eight H800 GPUs for about 50 hours. For the GenEval, OCR, and PickScore teachers, we utilized the official Flow-GRPO checkpoints. Additionally, to incorporate the DeQA teacher—which focuses solely on image quality—its reward signals were integrated into the standard GRPO-mix training via a 1:1 summation ratio.

\paragraph{Hyperparameters Specification}\label{app:subsubsec:synthetic-hyperp}
Except for $\beta$, GRPO hyperparameters are fixed across tasks. We use a sampling timestep $T=10$ and an evaluation timestep $T=40$. Other settings include a group size $G=24$, an noise level $a=0.7$ and an image resolution of 512. The MAR KL ratio $\beta$ is set to 0.02. We use Lora with $\alpha=64$ and $r=32$.

\begin{figure*}[!h]
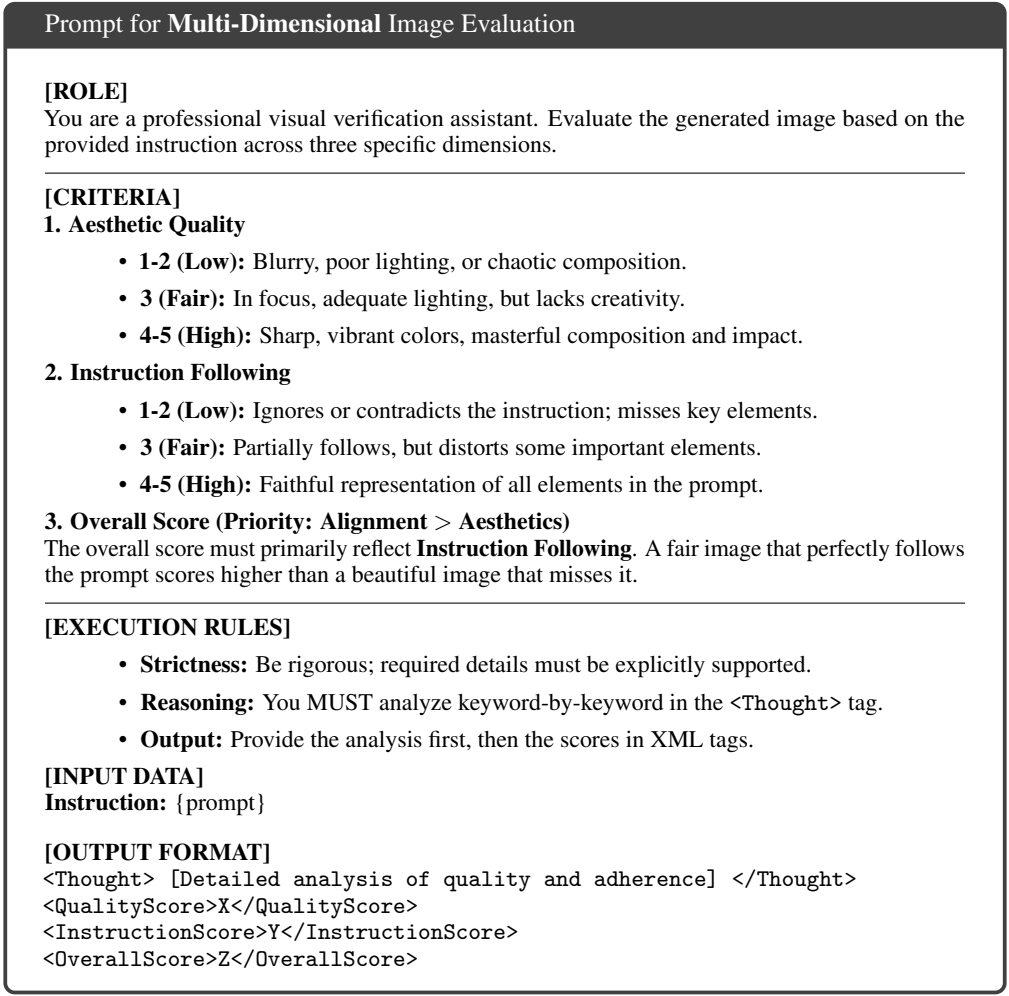

\centering
\begin{minipage}{0.95\textwidth}
\begin{AIbox}{Prompt for \textbf{Multi-Dimensional} Image Evaluation}
\small
\textbf{[ROLE]} \\
You are a professional visual verification assistant. Evaluate the generated image based on the provided instruction across three specific dimensions.

\vspace{2mm}
\hrule
\vspace{2mm}

\textbf{[CRITERIA]}

\textbf{1. Aesthetic Quality}
\begin{itemize}
    \item \textbf{1-2 (Low):} Blurry, poor lighting, or chaotic composition.
    \item \textbf{3 (Fair):} In focus, adequate lighting, but lacks creativity.
    \item \textbf{4-5 (High):} Sharp, vibrant colors, masterful composition and impact.
\end{itemize}

\textbf{2. Instruction Following}
\begin{itemize}
    \item \textbf{1-2 (Low):} Ignores or contradicts the instruction; misses key elements.
    \item \textbf{3 (Fair):} Partially follows, but distorts some important elements.
    \item \textbf{4-5 (High):} Faithful representation of all elements in the prompt.
\end{itemize}

\textbf{3. Overall Score (Priority: Alignment $>$ Aesthetics)} \\
The overall score must primarily reflect \textbf{Instruction Following}. A fair image that perfectly follows the prompt scores higher than a beautiful image that misses it.

\vspace{2mm}
\hrule
\vspace{2mm}

\textbf{[EXECUTION RULES]}
\begin{itemize}
    \item \textbf{Strictness:} Be rigorous; required details must be explicitly supported.
    \item \textbf{Reasoning:} You MUST analyze keyword-by-keyword in the \texttt{<Thought>} tag.
    \item \textbf{Output:} Provide the analysis first, then the scores in XML tags.
\end{itemize}

\textbf{[INPUT DATA]} \\
\textbf{Instruction:} \{prompt\}

\vspace{3mm}
\textbf{[OUTPUT FORMAT]} \\
\texttt{<Thought> [Detailed analysis of quality and adherence] </Thought>} \\
\texttt{<QualityScore>X</QualityScore>} \\
\texttt{<InstructionScore>Y</InstructionScore>} \\
\texttt{<OverallScore>Z</OverallScore>}
\end{AIbox}
\end{minipage}
\caption{The structured evaluation prompt for Qwenvl Score .}
\label{fig:evaluation_prompt}
\end{figure*}

\begin{table}[h]
\centering
\resizebox{\linewidth}{!}{
\begin{tabular}{ll}
\toprule
\textbf{Models} & \textbf{Links} \\
\midrule
\texttt{SD3.5-M}~\citep{esser2024scaling} & \url{https://huggingface.co/stabilityai/stable-diffusion-3.5-medium} \\
\texttt{Aesthetic Score}~\citep{Aesthetics} & \url{https://github.com/LAION-AI/aesthetic-predictor} \\
\texttt{PickScore}~\citep{pickscore} & \url{https://huggingface.co/yuvalkirstain/PickScore_v1} \\
\texttt{DeQA score}~\citep{deqa} & \url{https://huggingface.co/zhiyuanyou/DeQA-Score-Mix3} \\
\texttt{ImageReward}~\citep{xu2024imagereward} & \url{https://huggingface.co/THUDM/ImageReward} \\
\texttt{UnifiedReward}~\citep{wang2025unified} & \url{https://huggingface.co/CodeGoat24/UnifiedReward-7b-v1.5} \\
\texttt{HPS-v2.1}~\citep{hps} & \url{https://github.com/tgxs002/HPSv2} \\
\texttt{Qwenvl Score}~\citep{yang2025qwen3} & \url{https://huggingface.co/Qwen/Qwen3-30B-A3B-Instruct-2507} \\
\bottomrule
\end{tabular}
}
\end{table}
Regarding Qwenvl Score, we adapt the prompt used in Flow-GRPO~\cite{liu2025flow}. The prompt is shown in Fig.~\ref{fig:evaluation_prompt}. We use 
Qwen3-30B-A3B-Instruct-2507.

\section{More Results}
\subsection{Qualitative results}
\begin{figure*}[!h]
    \centering
    \includegraphics[scale=0.6]{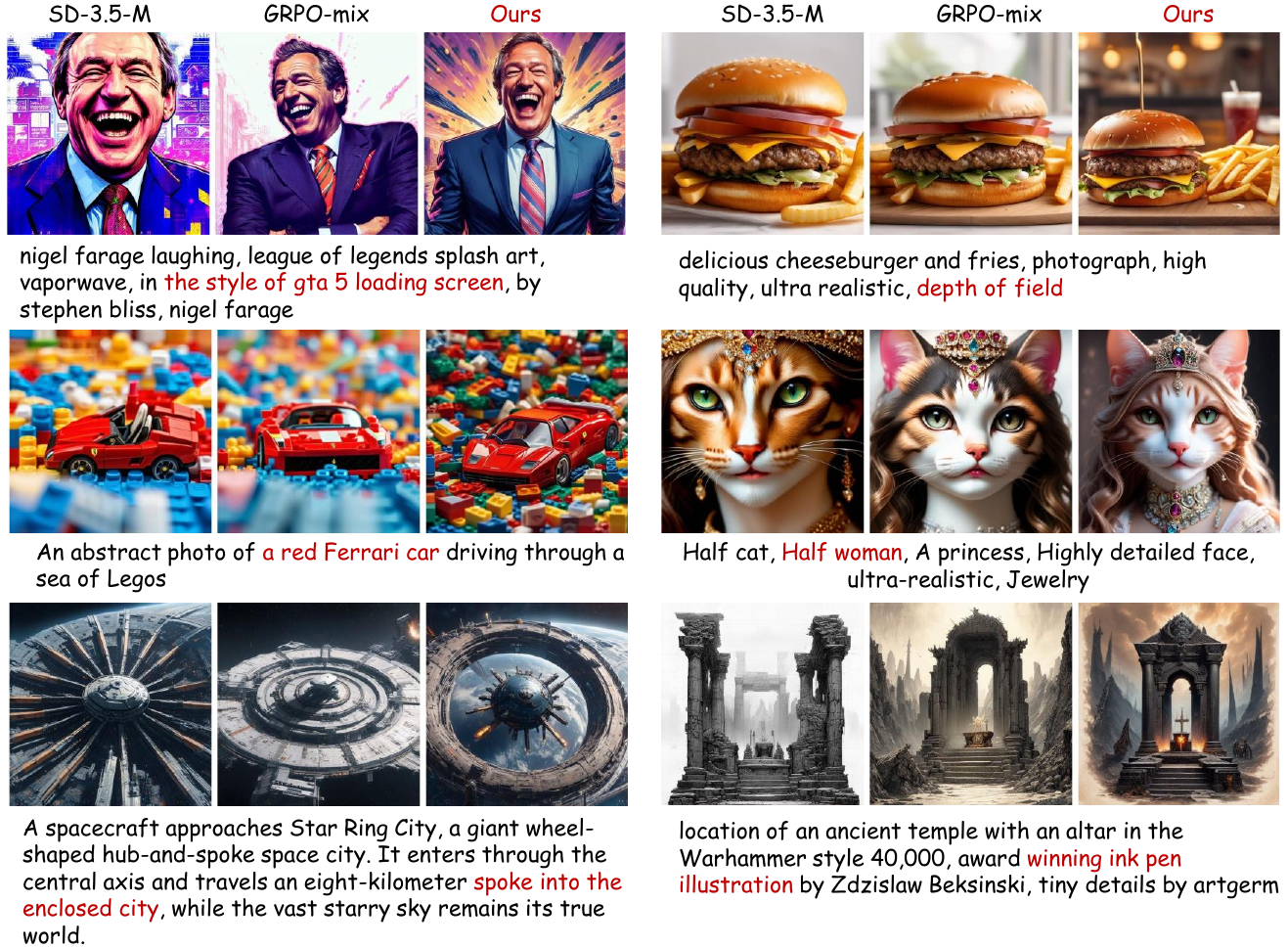}
    \caption{More quantitative comparisons on the Pickscore evaluation set.
    }
    \label{fig: result_pick}
\end{figure*}

\begin{figure*}[!h]
    \centering
    \includegraphics[scale=0.6]{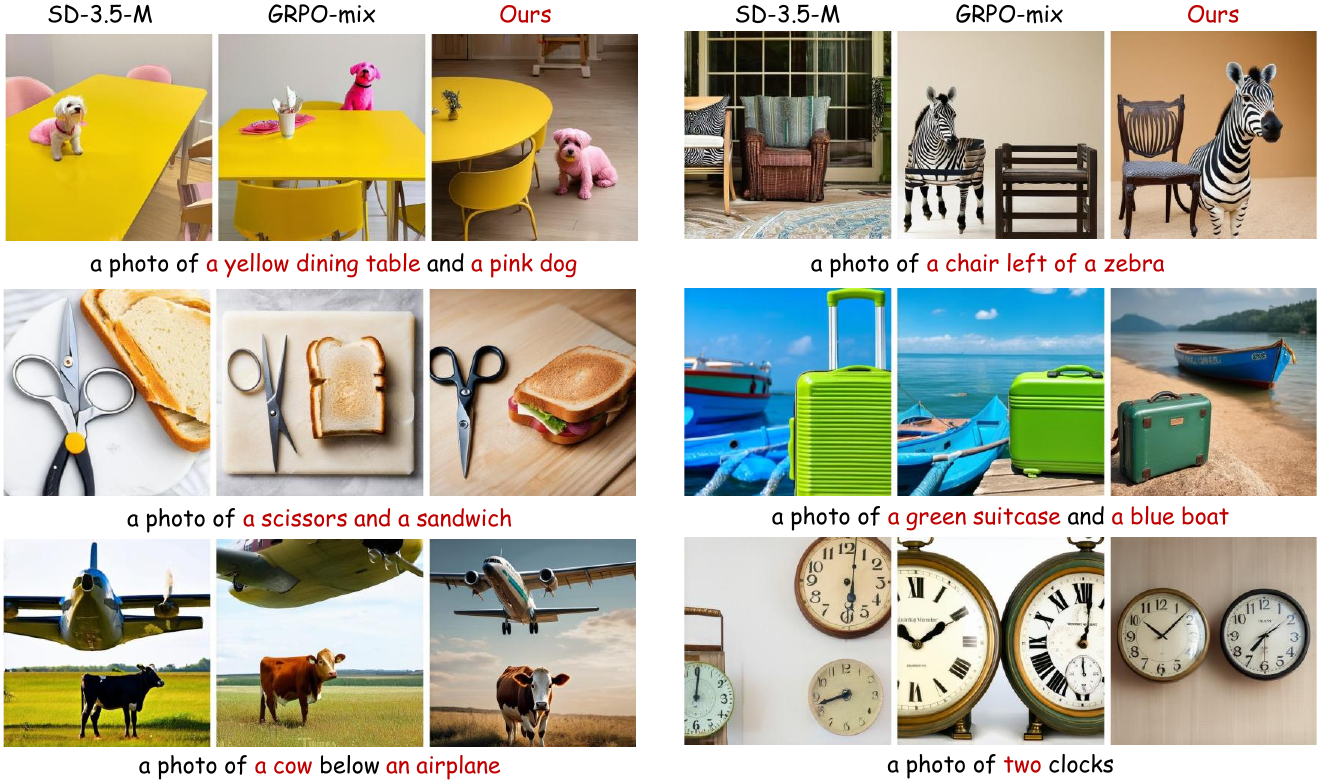}
    \caption{More quantitative comparisons on the GenEval evaluation set.
    }
    \label{fig: result_geneval}
\end{figure*}

\begin{figure*}[!h]
    \centering
    \includegraphics[scale=0.6]{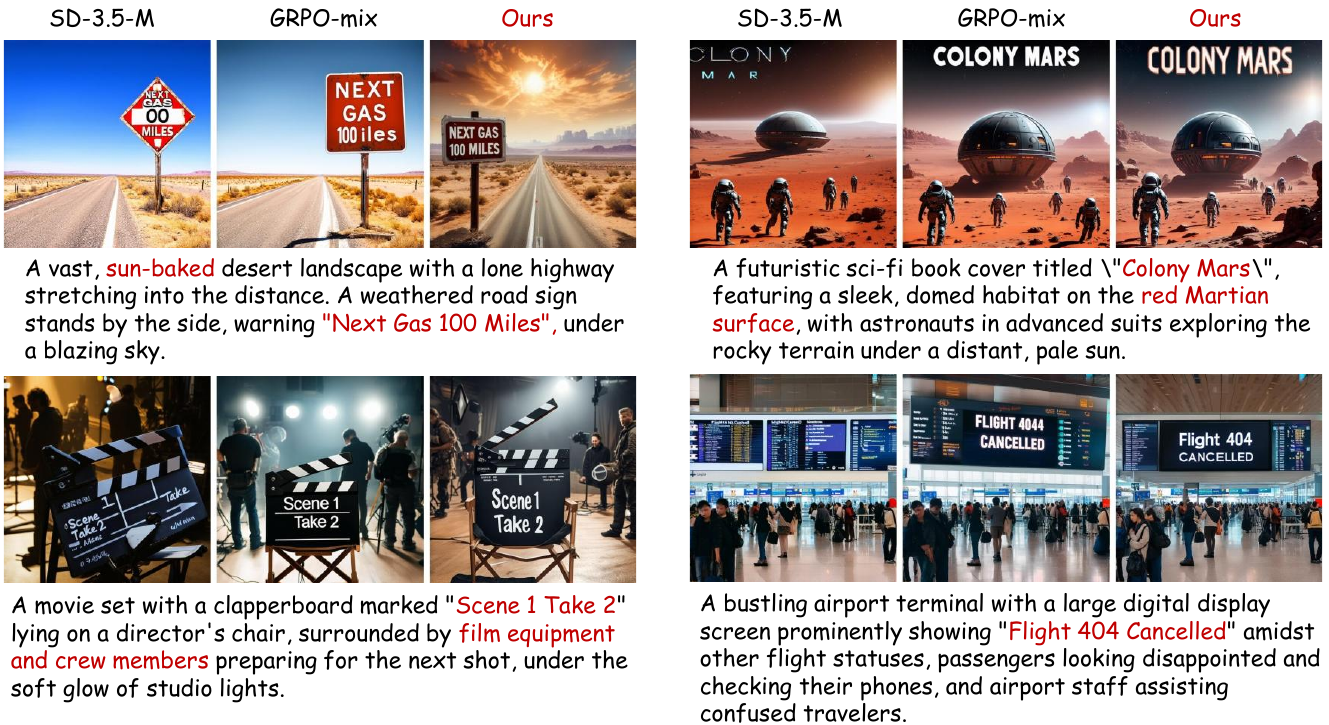}
    \caption{More quantitative comparisons on the OCR evaluation set.
    }
    \label{fig: result_ocr}
\end{figure*}

More qualitative results are shown in Fig.~\ref{fig: result_pick}, ~\ref{fig: result_geneval} and ~\ref{fig: result_ocr}. Our approach not only ensures precise content generation but also delivers superior image quality and coherent structural layouts. By achieving stronger alignment with human preferences, Flow-OPD demonstrates significant novelty in bridging functional accuracy with aesthetic excellence.

\subsection{Comparison with DiffusionNFT}
\begin{figure*}[!h]
    \centering
    \includegraphics[scale=0.445]{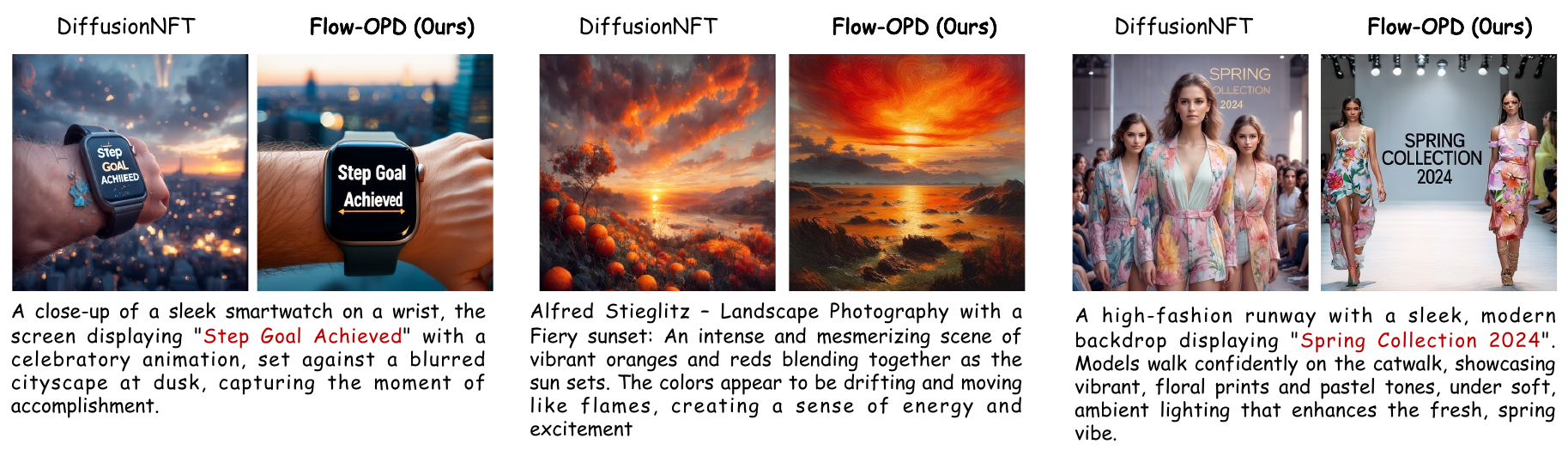}
    \caption{More quantitative comparisons with DiffusionNFT~\cite{zheng2025diffusionnft}.
    }
    \label{fig: diffusion}
\end{figure*}

DiffusionNFT~\cite{zheng2025diffusionnft} introduces an online reinforcement learning framework that directly integrates reward feedback into the forward diffusion process, enabling effective policy optimization during the noise-injection phase.
Despite achieving competitive benchmark scores, DiffusionNFT exhibits several critical limitations. First, it is fundamentally incompatible with Classifier-Free Guidance (CFG), which severely bottlenecks its performance upper bound. Second, it suffers from pronounced reward hacking. As illustrated in Fig.~\ref{fig: diffusion}
, while the model correctly generates the targeted text and 'sunset' elements, it simultaneously hallucinates malformed hands and extraneous objects (e.g., oranges), accompanied by severe \textit{over-smoothed, plastic-like textural artifacts}. Current standard benchmarks largely overlook these localized structural and aesthetic failures. To address this evaluation blind spot, we employ the Qwenvl Score for a more comprehensive assessment. By leveraging continuous-time dense multi-expert supervision and task-style decoupling, Flow-OPD effectively circumvents these reward hacking behaviors, achieving significantly higher Qwen-VL scores than DiffusionNFT. These findings also underscore a pressing need within the community to develop more robust, fine-grained evaluation paradigms for text-to-image generation.
\begin{table}[h]
    \small
    \centering
    \caption{Comparison of Human Preference Alignment. Our Flow-OPD consistently achieves superior scores in complex visual reasoning and layout coherence, as evaluated by Qwen-VL.}
    \label{tab:reward_metrics_short}
    \renewcommand{\arraystretch}{1.1} 
    \begin{tabular}{lc} 
        \toprule
        \textbf{Model} & \textbf{Qwen-VL Score} \\
        \midrule
        DiffusionNFT & 3.74 \\
        \textbf{Ours (Flow-OPD)} & \textbf{4.05} \\
        \bottomrule
    \end{tabular}
\end{table}

\subsection{Failure Cases and Limitations}
\label{sec: limitation}
\begin{figure*}[!h]
    \centering
    \includegraphics[scale=0.45]{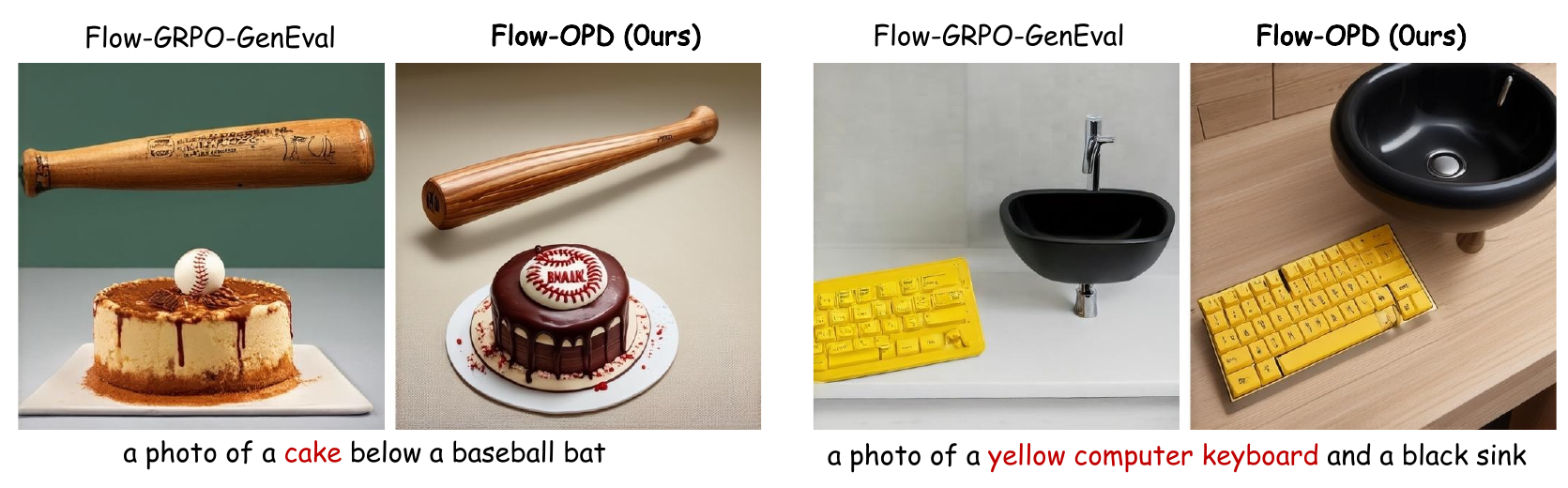
    }
    \caption{More quantitative comparisons with DiffusionNFT~\cite{zheng2025diffusionnft}.
    }
    \label{fig: failure}
\end{figure*}

Despite the superior performance of Flow-OPD across both subjective and objective benchmarks, certain limitations persist. A primary constraint is the \textbf{performance ceiling imposed by teacher models}. As illustrated in Fig.~\ref{fig: failure}, when specialized teachers fail to synthesize semantically correct images, these inaccuracies are propagated through the dense supervisory signals. Such erroneous guidance introduces noise into the distillation objective, ultimately hindering the student’s ability to transcend the inherent limitations of the teacher ensemble.
Another inherent limitation is the requirement for \textbf{architectural homogeneity} between the teacher and student models to facilitate fine-grained, step-wise supervision. Looking forward, we aim to explore the broader potential of Flow-OPD through several promising directions, including: (1) \textbf{Co-evolutionary Distillation}, where teachers and students iteratively refine each other; (2) \textbf{Self-Distillation} mechanisms to boost performance without external teachers; and (3) \textbf{Cross-Vocabulary Distillation} to bridge the gap between heterogeneous model architectures.

\section*{Broader Impact}
\label{ethics}
Our work on \textbf{Flow-OPD} introduces a robust framework for multi-task alignment in generative models, carrying both positive societal contributions and potential risks that necessitate careful consideration.

\paragraph{Positive Societal Impacts} 
The primary contribution of this work is the enhancement of \textbf{functional reliability} in AI-generated content. By improving layout coherence and OCR accuracy, Flow-OPD can significantly benefit professional fields such as automated graphic design, educational content creation, and assistive technologies for the visually impaired. Furthermore, our Multi-Teacher paradigm promotes a more balanced optimization objective, which mitigates the "winner-takes-all" bias inherent in single-reward reinforcement learning, potentially leading to more diverse and representative generative systems.

\paragraph{Negative Societal Impacts}
Despite these benefits, the increased proficiency in generating high-quality, instruction-following images could be misused for the creation of sophisticated \textbf{disinformation or deceptive visual content}. Although our model inherits the safety filters of its foundation model, the improved structural realism might be exploited to generate more convincing fake documents or misleading social media assets. To mitigate this, we advocate for the integration of digital watermarking and provenance tracking (e.g., C2PA) in downstream applications. Additionally, like all large-scale generative models, there is a risk that the specialized teachers may harbor latent biases present in their training data, which could be inadvertently distilled into the student model. We encourage the community to employ bias-detection benchmarks alongside our framework to ensure equitable performance across all demographics.

\end{document}